\pdfoutput=1
\documentclass[letterpaper]{article} % DO NOT CHANGE THIS
\usepackage{aaai25}  % DO NOT CHANGE THIS
\usepackage{times}  % DO NOT CHANGE THIS
\usepackage{helvet}  % DO NOT CHANGE THIS
\usepackage{courier}  % DO NOT CHANGE THIS
\usepackage[hyphens]{url}  % DO NOT CHANGE THIS
\usepackage{graphicx} % DO NOT CHANGE THIS
\usepackage{natbib}  % DO NOT CHANGE THIS AND DO NOT ADD ANY OPTIONS TO IT
\usepackage{caption} % DO NOT CHANGE THIS AND DO NOT ADD ANY OPTIONS TO IT
\usepackage{algorithm}
\usepackage{algorithmic}

\usepackage{newfloat}
\usepackage{listings}
\usepackage{amsmath}
\usepackage{amsthm}
\usepackage{amssymb}
\usepackage{tikz}
\usetikzlibrary{positioning, shapes.geometric, arrows.meta, calc}
\usepackage{subcaption}
\usetikzlibrary{decorations.markings}
\DeclareMathOperator*{\eqdef}{\stackrel{\text{\tiny def}}{=}}
\newtheorem{theorem}{Theorem}[section]
\newtheorem{lemma}[theorem]{Lemma}

\newtheorem{proposition}[theorem]{Proposition}
\DeclareMathOperator*{\argmax}{arg\,max}

\DeclareCaptionStyle{ruled}{labelfont=normalfont,labelsep=colon,strut=off} % DO NOT CHANGE THIS
\floatstyle{ruled}
\newfloat{listing}{tb}{lst}{}
\floatname{listing}{Listing}
\title{Howard's Policy Iteration is Subexponential for Deterministic Markov Decision Problems with Rewards of Fixed Bit-size and Arbitrary Discount Factor}

\author {
    Dibyangshu Mukherjee,
    Shivaram Kalyanakrishnan
}
\affiliations {
    Department of Computer Science and Engineering, IIT Bombay, Mumbai, India \\
    \{dbnshu, shivaram\}@cse.iitb.ac.in
}

\begin{document}

\maketitle

\begin{abstract}
Howard's Policy Iteration (HPI) is a classic algorithm for solving Markov Decision Problems (MDPs). HPI uses a ``greedy'' switching rule to update from any non-optimal policy to a dominating one, iterating until an optimal policy is found. Despite its introduction over $60$ years ago, the best-known upper bounds on HPI's running time remain exponential in the number of states---indeed even on the restricted class of MDPs with only deterministic transitions (DMDPs).
Meanwhile, the tightest lower bound for HPI for MDPs with a constant number of actions per state is only linear.
In this paper, we report a significant improvement:
a \textit{subexponential} upper bound for HPI on DMDPs, which is parameterised by the bit-size of the rewards, while independent of the discount factor. The same upper bound also applies to DMDPs with only two possible rewards (which may be of arbitrary size).
\end{abstract}
\section{Introduction}
Markov Decision Problems (MDPs)~\citep{bellman1967,puterman2014markov} are a well-studied abstraction of sequential decision making, which are widely used as a formal basis for automated control~\citep{Bertsekas:2007}, planning~\citep{kolobov2012planning} and reinforcement learning~\citep{Sutton+Barto:1998}. An MDP models an \textit{environment}, which comprises a set of states $S$ and a set of actions $A$. When action $a \in A$ is executed from state $s \in S$, the MDP transitions to a next state $s^{\prime} \in S$, while also emitting a numeric reward $r \in \mathbb{R}$. Transitions are in general stochastic: the probability of reaching $s^{\prime}$ by taking action $a$ from state $s$ is given by $T(s, a, s^{\prime})$, where $T: S \times A \times S \to [0, 1]$ (satisfying $\sum_{s^{\prime} \in S} T(s, a, s^{\prime}) = 1$ for $s \in S, a \in A$) is called the transition function of the MDP. Similarly, $r = R(s, a)$, where $R: S \times A \to \mathbb{R}$ is the reward function of the MDP. 

An \textit{agent} that interacts with an MDP goes through a sequence $(s^{t}, a^{t}, r^{t})_{t = 0}^{\infty}$, starting with initial state $s^{0} \in S$. The choice of which action $a^{t} \in A$
to execute at each time step $t \geq 0$ lies with the agent. This choice influences both the immediate reward $r^{t} = R(s^{t}, a^{t})$ and the next state $s^{t + 1} \sim T(s^{t}, a^{t}, \cdot)$. Hence, to maximise its \textit{long-term reward}, the agent's action choices must appropriately balance immediate and future rewards. Suppose that the agent fixes an action for each state---that is, it executes a \textit{policy} $\pi: S \to A$. In other words, the agent takes action $a^{t} = \pi(s^{t})$ for $t \geq 0$. For tasks with infinite horizons, there are two common definitions of long-term reward.

Under \textbf{discounted reward}, future rewards are discounted geometrically by a factor $\gamma \in [0, 1)$: the long-term reward (also called ``value'') of each state $s \in S$ under $\pi$ is 
\begin{align}
V_\gamma^\pi(s) = \mathbb{E}_\pi\left[  \sum_{t=0}^\infty \gamma^{t}R(s^t,a^t)      \bigg| \ s^0=s\right].
\label{eqn:discounted-value}    
\end{align}
Observe that the value depends on the discount factor $\gamma$, which is specified as a part of the MDP. 

An agent that follows policy $\pi$ also has a well-defined \textbf{average reward}, depending on its starting state $s \in S$. This long-term reward, called the \textit{gain} of $\pi$ from $s$, is given by 
\begin{align}
V^{\pi}_{\text{g}}(s) = \lim_{T \to \infty} \mathbb{E}_{\pi}\left[ \frac{1}{T}  \sum_{t  = 0}^{T - 1} R(s^{t}, a^{t}) \bigg| s^{0} = s\right].
\label{eqn:average-gain}    
\end{align}
Accompanying its gain is $\pi$'s \textit{bias}, defined for $s \in S$ as
\begin{align}
V^{\pi}_{\text{b}}(s) = \mathbb{E}_{\pi}\left[  \sum_{t  = 0}^{\infty} \left(R(s^{t}, a^{t})- V^{\pi}_{\text{g}}(s)\right) \bigg| s^{0} = s\right],
\label{eqn:average-bias}    
\end{align}
which is the cumulative difference over time between the reward and the average reward, starting from $s$.

For any given MDP, then, the natural problem is that of computing an \textit{optimal} policy. In the discounted reward setting,
it is established that for each input MDP $(S, A, T, R, \gamma)$, there exists an optimal policy $\pi^{\star}: S \to A$, which satisfies $V^{\pi^{\star}}_{\gamma}(s) \geq V^{\pi}_{\gamma}(s)$ for $s \in S$ and all $\pi: S \to A$. Similarly, under average reward,
each input MDP $(S, A, T, R)$ has an optimal policy $\pi^{\star}: S \to A$, which satisfies $V^{\pi^{\star}}_{\text{g}}(s) \geq V^{\pi}_{\text{g}}(s)$ for $s \in S$ and all $\pi: S \to A$, and which additionally satisfies certain ``optimality'' equations (provided in Appendix~\ref{app:averagereward}).

There are multiple algorithmic approaches to solve MDPs, including value iteration, linear programming, and policy iteration~\cite{Littman}. In this paper, we restrict our attention to Policy Iteration (PI)~\cite{howard1960dynamic,puterman2014markov}, which is widely used. PI is also conceptually simple: initialised at some arbitrary policy, at each iteration PI switches to a dominating policy, unless none exists. The running time of PI depends on the switching rule used for updating the policy. At one end of a spectrum are ``Simplex'' variants of PI that change the action at a single state, while the canonical greedy variant, called ``Howard's PI''~\cite{howard1960dynamic} (HPI), switches actions at \textit{all} ``improvable'' states. Yet other variants of PI~\cite{mansour1999complexity,kalyanakrishnan2016batch} switch some subset of improvable states. It is natural to wonder which variant is the ``most efficient''---and this question exposes a gap in our current theoretical understanding.

On the one hand, experiments have invariably shown HPI to significantly outperform other PI variants~\cite{kalyanakrishnan2016batch,taraviya2020tighter}. However, theoretical upper bounds on the iterations taken by HPI are still much looser than those for other variants. Consider finite MDPs with $|S| = n \geq 2$ states and $|A| = k \geq 2$ actions. Whereas upper bounds of $(O(\log k))^{n}$~\cite{kalyanakrishnan2016randomised} and $O(k^{0.7207n})$~\cite{gupta2017improved} have been shown for other variants, the tightest known upper bounds for HPI are $O(\frac{k^{n}}{n})$~\cite{mansour1999complexity}---only a linear improvement over the trivial bound of $k^{n}$. A similar trend plays out on Deterministic MDPs (DMDPs), the restricted class of MDPs in which every transition has a fixed next state: in other words, all transition probabilities are either $0$ or $1$. Whereas DMDPs can be solved in strongly polynomial time~\cite{karp1978characterization,madani2002}, and are in fact done so even by a Simplex variant of PI~\cite{post2013simplex}, the best upper bound for HPI on DMDPs remains exponential~\cite{goenka2025upper}.

Over the years, multiple researchers have explicitly earmarked improving the upper bound for HPI as an important target for theoretical research~\cite{Schmitz:1985,scherrer2013improved,post2013simplex}. This paper presents a significant step in this direction: a \textit{subexponential} upper bound for HPI on DMDPs, parameterised by the ``bit-size'' of the rewards.

\subsection{Bit-size of rewards}
\label{sec: bit-size}  
Consider DMDP $M = (S, A, T, R, \gamma)$ under discounted reward, or $M = (S, A, T, R)$ under average reward. Let $b^{\prime}(R)$ denote the smallest positive integer such that for each $(s, a) \in S \times A$, $R(s, a)$ belongs to the set $\{0, 1, 2, \dots, 2^{b^{\prime}(R)} - 1\}$ (with the  convention that $b^{\prime}(R) = \infty$ if no finite integer satisfies this requirement for $R$). Thus, if the only rewards in $R$ are $2$, $3$, and $5$, we would have $b^{\prime}(R) = 3$. We note that the sequence of policies visited by HPI on $M$ remains identical if the rewards in $M$ are positively scaled and/or shifted. Define, for $\alpha > 0, \beta \in \mathbb{R}$, and $(s, a) \in S \times A$: $R_{\alpha, \beta}(s, a) \eqdef \alpha R(s, a) + \beta$. We define the bit-size of $R$ by
\begin{align}
\label{eqn:b-def}
b(R) &\eqdef \min_{\alpha > 0, \beta \in \mathbb{R}} b^{\prime}(R_{\alpha, \beta}).
\end{align}
By this definition, $b(R)$ is $3$ not only for the set of rewards $\{2, 3, 5\}$, but among others, also for $\{0.2, 0.3, 0.5\}$, $\{2\sqrt{3}, 3\sqrt{3}, 5\sqrt{3}\}$, and $\{-0.8, -0.7, -0.5\}$. Note as a special case that $b(R) = 1$ for any reward function that takes exactly two values, since those values can be scaled and shifted to $\{0, 1\}$. In theory, $b(R)$ could be infinite (an example is when $1$, $\sqrt{2}$, and $\sqrt{3}$ are rewards). However, in applications of MDPs, rewards are invariably encoded in \texttt{float} or \texttt{double} variables, which are of fixed bit-size (say 32 or 64). In games such as Chess and Go, and
%several other
tasks where the only rewards are from success and failure, the reward set is typically $\{0, 1\}$ or $\{-1, 0, 1\}$~\citep{silver2016mastering,Silver2017MasteringCA}. In yet other tasks, reward is based on the number of time steps elapsed or energy expended, usually discretised to take a few tens or hundreds of contiguous integer values~\citep{crites1995improving}. It is therefore reasonable to consider $b(R)$ a \textit{constant} for tasks encountered in practice. Henceforth we shall drop ``$(R)$'' from ``$b(R)$'' when the context is clear. 

\subsection{Upper bounds and proof sketches}
\label{sec:UB-Pf-sketch}  
The effect of $b$ on the complexity of Policy Iteration is relatively straightforward to characterise under average reward. In this case, the gain-bias pair of every state, under every policy, is observed to belong to a finite set of size $\text{poly}(n, 4^{b})$. Regardless of which Policy Iteration variant is used, the strict monotonicity of the algorithm yields the following upper bound.

\begin{proposition}
\label{prop:averagereward}
Let $M = (S, A, T, R)$ be a DMDP using average reward, with $|S| = n$ and $|A| = k$. On $M$, any Policy Iteration algorithm can visit at most \( O\left( n^5 \! \cdot \! 4^{b} \right) \) policies.
\end{proposition}
A proof of the proposition is given in Appendix~\ref{app:averagereward}. The pseudopolynomial dependence on the rewards is a novel result for HPI. However, the simple argument behind this upper bound for average reward does not carry over to discounted reward. With discounted reward, the number of possible values a state could  have (under different policies) scales exponentially, rather than polynomially, in $n$. See section~\ref{sec:Avg-vs-Disc}. Our main result is that the number of iterations taken by HPI still only scales \textit{subexponentially} in $n$ if $b$ is a constant, or even if $b = o(n^{1 - \epsilon})$ for some $\epsilon > 0$. 
\begin{theorem}
\label{thm:discountedreward}
Let $M = (S, A, T, R, \gamma)$ be a DMDP using discounted reward, with $|S| = n$ and $|A| = k$. On $M$, HPI can visit at most $n k \cdot u \exp(u)$ policies, where $u = O\left(\sqrt{nb} \log \frac{n}{b} + b \right)$.
\end{theorem}

The trajectory followed by HPI on the input MDP depends on the discount factor $\gamma$. First, we show that there exists a threshold discount factor, denoted $\gamma_{\text{Q}}$, beyond which the trajectory of HPI is invariant. In other words, for two MDPs $M_\gamma$ and $M_{\gamma'}$ that differ only in their discount factors $\gamma$ and $\gamma'$, HPI will follow the same trajectory as long as $\gamma, \gamma' \in (\gamma_{\text{Q}}, 1)$. An upper bound parameterised by $\gamma$, of $O\left(\frac{n k}{1 - \gamma} \log \frac{1}{1 - \gamma} \right)$ iterations, is already known for HPI~\cite{ye2011simplex,scherrer2013improved}. For $\gamma > \gamma_{\text{Q}}$, the invariance of the algorithm's trajectory therefore implies an upper bound of $O\left(\frac{n k}{1 - \gamma_{\text{Q}}} \log \frac{1}{1 - \gamma_{\text{Q}}} \right)$ iterations. Our second innovation is to obtain the subexponential-in-$n$ upper bound on $\frac{1}{1 - \gamma_{\text{Q}}}$. This is achieved by showing that $\gamma_{\text{Q}}$ is the largest root to the left of $1$ of a certain polynomial with integer coefficients with magnitude $O(2^{b})$. We combine some existing results on root-separation in such polynomials to obtain our upper bound.

The average reward setting provides useful intuitions to reason about discounted rewards. We draw connections between these settings where appropriate, but defer the detailed analysis under average reward to Appendix~\ref{app:averagereward}. The main text remains focused on discounted rewards; the proof of Theorem~\ref{thm:discountedreward} is given in Section~\ref{sec:Disc-reward-UB}. We proceed to formal definitions of PI and DMDPs (Section~\ref{sec:background}), followed by a survey of related work (Section~\ref{sec:related-work}).

\section{Background}
\label{sec:background}

In this section, we provide the necessary background on Policy Iteration under discounted reward, and also on DMDPs, thus laying the foundation for our main result in Section~\ref{sec:Disc-reward-UB}.
\subsection{Policy Iteration}
\label{sec:PI}
Policy Iteration (PI) builds upon the idea that if a policy $\pi: S \to A$ is non-optimal, then it is possible to efficiently identify a set of policies that are guaranteed to improve upon $\pi$ (in terms of long-term reward). Any policy $\pi^{\prime}: S \to A$ from this improving set is then made the current policy, and the process continued. Since there are only a finite number of policies ($k^{n}$), and no policy can repeat due to the monotonic improvement, the process is guaranteed to terminate at an optimal policy after a finite number of iterations. 

\subsubsection{Policy evaluation}
The first step in each iteration of PI is \textit{policy evaluation}: that is, to compute the value of $\pi$.
State values as defined in \eqref{eqn:discounted-value} can be computed as the solution of a system of linear equations called the Bellman equations~\citep{bellman1967}: for $s \in S$,
\begin{align}
\label{eqn:disocunted-bellman}
V_{\gamma}^\pi(s) &= R(s,\pi(s)) + \gamma \sum_{s' \in S}{T\left(s,\pi(s),s'\right) V_{\gamma}^\pi(s')}.
\end{align}

\subsubsection{Policy improvement}
The basis for policy improvement is to consider the effect of taking an alternative action $a \in A$ from $s$ only for the first time step, and thereafter acting according to $\pi$ (at time steps $2, 3, 4, \dots$). These long-term rewards are called Q-values, and are defined as follows for each $(s, a) \in S \times A$:
\begin{align}
\label{eqn:q-discounted-mdp}
Q_{\gamma}^\pi(s,a) &= R(s,a) + \gamma \sum_{s' \in S}{T \left(s,a,s' \right)V_{\gamma}^\pi(s')}.
\end{align}

The intuition behind policy improvement is that if it is more rewarding to take action $a$ from state $s$ just once and then follow $\pi$ (the long-term reward for this is precisely the Q-value), then it must also be beneficial to take $a$ from $s$ all the time---that is, whenever the agent finds itself in $s$. The latter would amount to following a new policy that takes $a$ from $s$, but which mimics $\pi$ at all other states. Continuing with the same intuition, one could similarly switch to actions with higher Q-values simultaneously in any number of states where such actions are available. The ``policy improvement theorem'', given below, formalises this intuition.
\begin{theorem}[Policy Improvement]
\label{thm:pit-discounted}    
Consider MDP $M = (S, A, T, R, \gamma)$ using discounted reward.
\begin{enumerate}
\item Suppose policy $\pi: S \to A$ satisfies $Q^{\pi}_{\gamma}(s, a) \leq V^{\pi}_{\gamma}(s)$ for all $s \in S, a \in A$. Then $\pi$ is an optimal policy.
\item Suppose policies $\pi: S \to A$ and $\pi^{\prime}: S \to A$ satisfy
$Q^{\pi}_{\gamma}(s, \pi^{\prime}(s)) \geq V^{\pi}_{\gamma}(s)$ for all $s \in S$, and moreover, there exists $s \in S$ such that $Q^{\pi}_{\gamma}(s, \pi^{\prime}(s)) > V^{\pi}_{\gamma}(s).$ Then $V^{\pi^{\prime}}_{\gamma}(s) \geq V^{\pi}_{\gamma}(s)$ for all $s \in S$, and there exists $s \in S$ such that $V^{\pi^{\prime}}_{\gamma}(s) > V^{\pi}_{\gamma}(s)$.
\end{enumerate}
\end{theorem}
The proof of Theorem~\ref{thm:pit-discounted} relies on defining an operator that becomes a contraction mapping in a Banach space, on account of the discount factor $\gamma$ being smaller than $1$~\citep{Bertsekas:2007,Szevespari}. 

\subsection{Howard's PI}
\label{sec:HPI}

Notice from Theorem~\ref{thm:pit-discounted} that once $\pi$ is evaluated, in general there could be multiple choices of improving policies $\pi^{\prime}$. Any action $a \in A$ that has a higher Q-value for state $s \in S$ is called an ``improving action'', and $s$ itself is an ``improvable state''. As long as we switch to some improving action in one or more states, and we retain the actions of $\pi$ at the other states, the resulting policy $\pi^{\prime}$ is guaranteed to strictly dominate $\pi$. HPI is a greedy variant of PI in which \textit{every} state that has some improving action is necessarily switched. If there are multiple improving actions, any one with the highest Q-value is selected. Below is a full specification of HPI.

\begin{center}
\framebox[0.45\textwidth]{
\parbox{0.4\textwidth}{
\underline{HPI starting from policy $\pi$ under discounted reward}\\\\
1. For $s \in S$: set $\pi^{\prime}(s) \gets \argmax_{a \in A} Q^{\pi}_{\gamma}(s, a)$, breaking ties in favour of $\pi(s)$ if possible, else arbitrarily.\\ 
2. If $\pi^{\prime} \neq \pi$, set $\pi \gets \pi^{\prime}$ and go to 1.\\
3. Declare $\pi$ to be optimal and terminate.
}
}
\end{center}

% If each reward in $R$ is scaled by $\alpha > 0$ and shifted by $\beta \in \mathbb{R}$, the same happens to each Q-value. 
If each reward in \(R\) is scaled by \(\alpha > 0\) and shifted by \(\beta \in \mathbb{R}\), then each Q-value is scaled by \(\alpha\) and shifted by \(\frac{\beta}{1-\gamma}\). 
From the code above, it is clear that positive-scaling and shifting have no effect on the trajectory taken by HPI (except possibly for tie-breaking). This invariance validates our definition of $b(R)$.

\subsection{DMDPs}
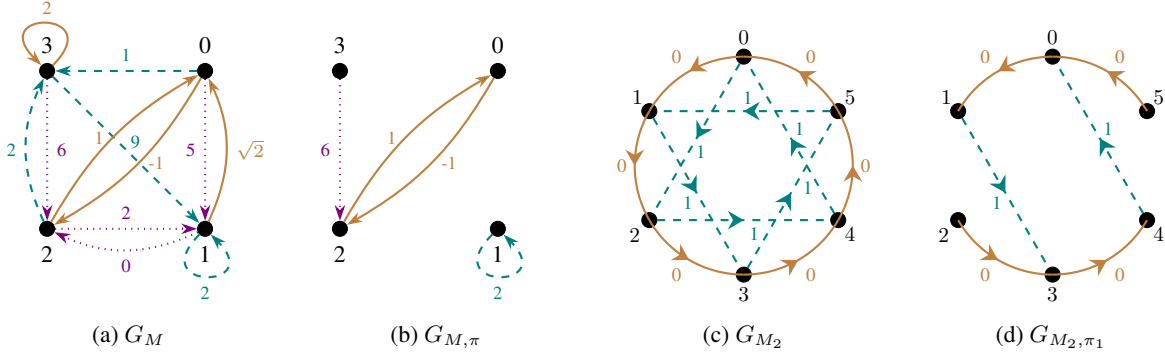
\begin{figure*}[t]
  \centering
  \begin{subfigure}{0.225\linewidth} 
    \centering
    \begin{tikzpicture}[scale=1.75, 
      >={Stealth[length=6pt,width=4pt]},
      every edge/.append style={thick},
      vertex/.style={draw, fill=black, circle, minimum size=5pt, inner sep=2pt},
      brown edge/.style={draw=brown},
      teal edge/.style={draw=teal},
      violet edge/.style={draw=violet}
      ]
      % vertices
      \node[vertex,label={above:\footnotesize 3}] (D) at (0,0) {};
      \node[vertex,label={above:\footnotesize 0}] (A) at (1.2,0) {};
      \node[vertex,label={below:\footnotesize 1}] (B) at (1.2,-1.2) {};
      \node[vertex,label={below:\footnotesize 2}] (C) at (0,-1.2) {};

      % edges (violet)
      \path[->, dotted, violet edge] (A) edge node[midway, left, violet] {\scriptsize 5} (B);
      \path[->, dotted, violet edge] (C) edge node[midway, above, violet] {\scriptsize 2} (B);
      \path[->, dotted, violet edge] (D) edge node[midway, right, violet] {\scriptsize 6} (C);
      \path[->, dotted, violet edge] (B) edge[bend left=25] node[midway, below, violet] {\scriptsize 0} (C);

      % edges (teal)
      \path[->, dashed, teal edge] (A) edge node[midway, above, teal] {\scriptsize 1} (D);
      \path[->, dashed, teal edge] (D) edge node[midway, right, teal,xshift=-2pt,yshift=3pt] {\scriptsize 9} (B);
      \path[->, dashed, teal edge] (C) edge[bend left=25] node[midway, left, teal] {\scriptsize 2} (D);
      \path[->, dashed, teal edge, loop, out=225, in=315, min distance=6mm] (B) edge node[below, teal] {\scriptsize 2} (B);

       % edges (brown)
      \path[->, brown edge] (A) edge[bend left=15] node[midway, right, brown] {\scriptsize -1} (C);
      \path[->, brown edge] (C) edge[bend left=15] node[midway, left, brown] {\scriptsize 1} (A);
      \path[->, brown edge] (B) edge[bend right=25] node[midway, right, brown] {\scriptsize $\sqrt{2}$} (A);
      \path[->, brown edge, loop, out=45, in=135, min distance=6mm] (D) edge node[above, brown] {\scriptsize 2} (D);

    \end{tikzpicture}
    \caption{$G_{M}$}
    \label{fig:a}
  \end{subfigure}
  \begin{subfigure}{0.225\linewidth} 
    \centering
    \begin{tikzpicture}[scale=1.75, 
      >={Stealth[length=6pt,width=4pt]},
      every edge/.append style={thick},
      vertex/.style={draw, fill=black, circle, minimum size=5pt, inner sep=2pt},
      brown edge/.style={draw=brown},
      teal edge/.style={draw=teal},
      violet edge/.style={draw=violet}
      ]
      % vertices
      \node[vertex,label={above:\footnotesize 3}] (D) at (0,0) {};
      \node[vertex,label={above:\footnotesize 0}] (A) at (1.2,0) {};
      \node[vertex,label={below:\footnotesize 1}] (B) at (1.2,-1.2) {};
      \node[vertex,label={below:\footnotesize 2}] (C) at (0,-1.2) {};

      \path[->, dotted, violet edge] (D) edge node[midway, left, violet] {\scriptsize 6} (C);

      \path[->, dashed, teal edge, loop, out=225, in=315, min distance=6mm] (B) edge node[below, teal] {\scriptsize 2} (B);

       % edges (brown)
      \path[->, brown edge] (A) edge[bend left=15] node[midway, right, brown] {\scriptsize -1} (C);
      \path[->, brown edge] (C) edge[bend left=15] node[midway, left, brown] {\scriptsize 1} (A);

    \end{tikzpicture}
    \caption{$G_{M,\pi}$}
    \label{fig:b}
  \end{subfigure}
    \begin{subfigure}{0.225\linewidth}
        \centering
        \begin{tikzpicture}[scale=0.29, every node/.style={scale=0.8}]
            % Define the radius of the circle
            \def\radius{5cm}
            % Define the arrow size
            \tikzset{>={Stealth[length=6pt,width=7pt]}}
        
            % Manually place the nodes and labels
            \node[draw, circle, fill=black, inner sep=2.5pt, label={[label distance=-0.4mm]90:$0$}] (A) at (90:\radius) {};
            \node[draw, circle, fill=black, inner sep=2.5pt, label={[label distance=-1.5mm]150:$1$}] (B) at (150:\radius) {};
            \node[draw, circle, fill=black, inner sep=2.5pt, label={[label distance=-1mm]210:$2$}] (C) at (210:\radius) {};
            \node[draw, circle, fill=black, inner sep=2.5pt, label={[label distance=-0.5mm]270:$3$}] (D) at (270:\radius) {};
            \node[draw, circle, fill=black, inner sep=2.5pt, label={[label distance=-1.5mm]330:$4$}] (E) at (330:\radius) {};
            \node[draw, circle, fill=black, inner sep=2.5pt, label={[label distance=-1.5mm]30:$5$}] (F) at (30:\radius) {};

            % Draw the circular contour with arrows and labels between the nodes in brown
            \draw[brown, thick, postaction={decorate, decoration={markings, mark=at position 0.5 with {\arrow{>}}}}] 
                (A) arc (90:150:\radius) node[midway, above left, font=\footnotesize] {0};
            \draw[brown, thick, postaction={decorate, decoration={markings, mark=at position 0.5 with {\arrow{>}}}}] 
                (B) arc (150:210:\radius) node[midway, left, font=\footnotesize] {0};
            \draw[brown, thick, postaction={decorate, decoration={markings, mark=at position 0.5 with {\arrow{>}}}}] 
                (C) arc (210:270:\radius) node[midway, below left, font=\footnotesize] {0};
            \draw[brown, thick, postaction={decorate, decoration={markings, mark=at position 0.5 with {\arrow{>}}}}] 
                (D) arc (270:330:\radius) node[midway, below right, font=\footnotesize] {0};
            \draw[brown, thick, postaction={decorate, decoration={markings, mark=at position 0.5 with {\arrow{>}}}}] 
                (E) arc (330:390:\radius) node[midway, right, font=\footnotesize] {0};
            \draw[brown, thick, postaction={decorate, decoration={markings, mark=at position 0.5 with {\arrow{>}}}}] 
                (F) arc (390:450:\radius) node[midway, above right, font=\footnotesize] {0};

            % Draw the internal oriented star edges with labels, make them teal
            \draw[teal, dashed, thick, postaction={decorate, decoration={markings, mark=at position 0.5 with {\arrow{>}}}}] 
                (A) -- node[auto, above, fill=white, xshift=2pt, yshift=-10pt, inner sep=0.5pt, outer sep=0pt, font=\footnotesize] {1} (C);
            \draw[teal, dashed,thick, postaction={decorate, decoration={markings, mark=at position 0.5 with {\arrow{>}}}}] 
                (B) -- node[auto, above, fill=white, xshift=-4pt, yshift=-8pt, inner sep=0.5pt, outer sep=0pt, font=\footnotesize] {1} (D);
            \draw[teal, dashed, dashed, thick, postaction={decorate, decoration={markings, mark=at position 0.5 with {\arrow{>}}}}] 
                (C) -- node[auto, above, fill=white, xshift=5pt, yshift=-10pt, inner sep=0.5pt, outer sep=0pt, font=\footnotesize] {1} (E);
            \draw[teal, dashed,thick, postaction={decorate, decoration={markings, mark=at position 0.5 with {\arrow{>}}}}] 
                (D) -- node[auto, above, fill=white, xshift=8pt, yshift=-1pt, inner sep=0.5pt, outer sep=0pt, font=\footnotesize] {1} (F);
            \draw[teal, dashed, thick, postaction={decorate, decoration={markings, mark=at position 0.5 with {\arrow{>}}}}] 
                (E) -- node[auto, above, fill=white, xshift=4pt, yshift=1pt, inner sep=0.5pt, outer sep=0pt, font=\footnotesize] {1} (A);
            \draw[teal, dashed, thick, postaction={decorate, decoration={markings, mark=at position 0.5 with {\arrow{>}}}}] 
                (F) -- node[auto, above, fill=white, xshift=2pt, yshift=3pt, inner sep=0.5pt, outer sep=0pt, font=\footnotesize] {1} (B);
    
        \end{tikzpicture}
        \caption{$G_{M_{2}}$}
        \label{fig:c}
      \end{subfigure}
      \begin{subfigure}{0.225\linewidth}
        \centering
        \begin{tikzpicture}[scale=0.29, every node/.style={scale=0.8}]
            % Define the radius of the circle
            \def\radius{5cm}
            % Define the arrow size
            \tikzset{>={Stealth[length=6pt,width=7pt]}}
        
            % Manually place the nodes and labels
            \node[draw, circle, fill=black, inner sep=2.5pt, label={[label distance=-0.4mm]90:$0$}] (A) at (90:\radius) {};
            \node[draw, circle, fill=black, inner sep=2.5pt, label={[label distance=-1.5mm]150:$1$}] (B) at (150:\radius) {};
            \node[draw, circle, fill=black, inner sep=2.5pt, label={[label distance=-1mm]210:$2$}] (C) at (210:\radius) {};
            \node[draw, circle, fill=black, inner sep=2.5pt, label={[label distance=-0.5mm]270:$3$}] (D) at (270:\radius) {};
            \node[draw, circle, fill=black, inner sep=2.5pt, label={[label distance=-1.5mm]330:$4$}] (E) at (330:\radius) {};
            \node[draw, circle, fill=black, inner sep=2.5pt, label={[label distance=-1.5mm]30:$5$}] (F) at (30:\radius) {};

            % Draw the circular contour with arrows and labels between the nodes in brown
            \draw[brown, thick, postaction={decorate, decoration={markings, mark=at position 0.5 with {\arrow{>}}}}] 
                (A) arc (90:150:\radius) node[midway, above left, font=\footnotesize] {0};
            \draw[brown, thick, postaction={decorate, decoration={markings, mark=at position 0.5 with {\arrow{>}}}}] 
                (C) arc (210:270:\radius) node[midway, below left, font=\footnotesize] {0};
            \draw[brown, thick, postaction={decorate, decoration={markings, mark=at position 0.5 with {\arrow{>}}}}] 
                (D) arc (270:330:\radius) node[midway, below right, font=\footnotesize] {0};
            \draw[brown, thick, postaction={decorate, decoration={markings, mark=at position 0.5 with {\arrow{>}}}}] 
                (F) arc (390:450:\radius) node[midway, above right, font=\footnotesize] {0};

            \draw[teal, dashed,thick, postaction={decorate, decoration={markings, mark=at position 0.5 with {\arrow{>}}}}] 
                (B) -- node[auto, above, fill=white, xshift=-4pt, yshift=-8pt, inner sep=0.5pt, outer sep=0pt, font=\footnotesize] {1} (D);
            \draw[teal, dashed, thick, postaction={decorate, decoration={markings, mark=at position 0.5 with {\arrow{>}}}}] 
                (E) -- node[auto, above, fill=white, xshift=4pt, yshift=1pt, inner sep=0.5pt, outer sep=0pt, font=\footnotesize] {1} (A);

        \end{tikzpicture}
        \caption{$G_{M_{2}, \pi_{1}}$}
        \label{fig:d}
      \end{subfigure}

\caption{In (a) we see graph $G_{M}$ induced by DMDP $M$ with $S = \{0, 1, 2, 3\}$, $A = \{0 \text{ (solid)}, 1 \text{ (dashed)},  2 \text{ (dotted)}\}$. Each transition is annotated with the reward. Subfigure (b) shows the subgraph $G_{M, \pi}$ obtained by fixing policy $\pi$ such that $\pi(0) = 0, \pi(1) = 1, \pi(2) = 0, \pi(3) = 2$. Starting from each state $s$ in $G_{M, \pi}$ is a (possibly null) path $P^{\pi}_{s}$ and a cycle $C^{\pi}_{s}$ (defined in Section~\ref{sec:path-cycle} as a sequence of state-action pairs). For example: $P^{\pi}_{1} = \emptyset, C^{\pi}_{1} = \langle (1, 1) \rangle$; $P^{\pi}_{2} = \emptyset, C^{\pi}_{2} = \langle (2, 0),(0,0) \rangle$; and $P^{\pi}_{3} = \langle (3, 2) \rangle, C^{\pi}_{3} = \langle (2, 0),(0,0) \rangle$. In (c) we see $G_{M_{m}}$ for $m = 2$, containing 6 states and 2 actions. The family $M_{m}$ is described in Section~\ref{sec:Avg-vs-Disc}. Subfigure (d) shows the subgraph of $G_{M}$ induced by some fixed policy $\pi_{1}$. Notice that state $0$ is on a cycle with $4$ edges, exactly $2$ of which have a reward of $1$.
}
\label{fig}
\end{figure*}

In a DMDP $M = (S, A, T, R, \gamma)$, transition probabilities are all $0$ or $1$. Hence for $s \in S, a \in A$, it becomes convenient to denote as $T(s, a)$ the single next state $s^{\prime}$ for which $T(s, a, s^{\prime}) = 1$. DMDP $M$ induces the multi-graph $G_{M} = (\mathcal{V}, \mathcal{E}, w)$, which encodes the states of $M$ as its vertices, the transitions as edges, and the rewards as edge weights. Formally, (1) $\mathcal{V} = S$; (2) for $v_{1}, v_{2} \in \mathcal{V}$ and $a \in A$, $(v_{1}, v_{2}, a) \in \mathcal{E}$ iff $T(v_{1}, a) = v_{2}$; and (3) $w: \mathcal{E} \to \mathbb{R}$ is such that for $v_{1}, v_{2} \in \mathcal{V}$ and $a \in A$, $w((v_{1}, v_{2}, a)) = R(v_{1}, a)$. Figure~\ref{fig:a} shows an example of the induced graph. 

\subsubsection{Paths and cycles induced by DMDPs.}
\label{sec:path-cycle}
To study any fixed policy $\pi: S \to A$, we may consider the subgraph $G_{M, \pi}$ obtained by dropping from $G_{M}$ all the edges \textit{not} corresponding to actions taken by $\pi$. Thus, $G_{M, \pi}$ has exactly one edge---$(s, T(s, \pi(s)), \pi(s))$---for each state $s \in S$. For each $s \in S$, observe that starting from $s$ and repeatedly taking actions according to $\pi$ yields a (possibly empty) path, which is followed by a cycle. Let $C^{\pi}_{s}$ be the sequence of state-action pairs in the cycle that is reached from $s$ by following $\pi$, beginning with the first state in the cycle that is thereby reached. Let $c^{\pi}_{s}$ be the length of $C^{\pi}_{s}$. Let $P^{\pi}_{s}$ be the sequence of state-action pairs in the path emanating from $s$ under $\pi$, with its final element being the state-action pair that reaches the starting state of $C^{\pi}_{s}$. Let $p^{\pi}_{s}$ be the length of $P^{\pi}_{s}$. Figure~\ref{fig:a} provides some examples of $P^{\pi}_{s}$ and $C^{\pi}_{s}$ along with the DMDP shown for illustration. Observe that in general, $0 \leq p^{\pi}_{s} \leq n - 1$ and $1 \leq c^{\pi}_{s} \leq n$ for all $\pi: S \to A$ and $s \in S$. For $1 \leq i \leq p^{\pi}_{s}$, we denote by $P^{\pi}_{s}[i]$ the $i$-th element of $P^{\pi}_{s}$. Similarly, for $1 \leq i \leq c^{\pi}_{s}$, we denote by $C^{\pi}_{s}[i]$ the $i$-th element of $C^{\pi}_{s}$. If, say, this element is $(\bar{s}, \bar{a}) \in S \times A$, then $R(C^{\pi}_{s}[i])$ shall denote $R(\bar{s}, \bar{a})$.

With this notation, we obtain for DMDP $(S, A, T, R, \gamma)$ using discounted reward, that for $s \in S$,
\begin{align}
V^{\pi}_{\gamma}(s) &= \sum_{i = 1}^{p^{\pi}_{s}} \gamma^{i - 1} R(P^{\pi}_{s}[i]) + \nonumber\\ &~~~~~~~~\frac{\gamma^{p^{\pi}_{s}}}{1 - \gamma^{c^{\pi}_{s}}} \sum_{i = 1}^{c^{\pi}_{s}} \gamma^{i - 1} R(C^{\pi}_{s}[i]).
\label{eqn:discounted-raw}
\end{align}
Also, simplifying \eqref{eqn:q-discounted-mdp} for DMDPs yields for $s \in S, a \in A$: 
\begin{equation}\label{eqn:q-discounted-dmdp}
    Q_{\gamma}^\pi(s,a) = R(s,a) + \gamma V_{\gamma}^\pi(T(s, a)). 
\end{equation}

\subsection{Average reward versus discounted reward}
\label{sec:Avg-vs-Disc}

Before proceeding, we describe a technical difference between the average reward and discounted reward settings, which illustrates why the latter is more challenging for PI. 

Consider the family of DMDPs \( M_m \), where for \( m \geq 1 \), we have \( n = 3m \) and \( k = 2 \). For each state $s \in S \eqdef \{0, 1, \dots, n - 1\}$, action $0$ earns a reward of $0$ and action $1$ a reward of $1$. Also, for $s \in S$, $T(s, 0) = (s+1) \mod n$, while $T(s, 1) = (s+2) \mod n$.
Figure~\ref{fig:c} shows the graph induced by $M_2$.

Now, let \( S_{2m} \) denote the set of bit-strings of length $2m$ that contain an equal number of 0’s and 1’s. Clearly, \( |S_{2m}| = \binom{2m}{m} = \binom{2n/3}{n/3} = \exp(\Omega(n)) \). We observe that each bit-string in \( S_{2m} \) can be mapped to a cycle of length \( 2m \) in $M_{m}$, which passes through state $0$, and with a sequence of rewards the same as in the bit-string. For example, Figure~\ref{fig:d} shows a cycle corresponding to bit-string $0101$. In the average reward setting, all such cycles will have the \textit{same} gain, of \( \frac{1}{2} \). However, the (discounted) value of state \( 0 \) in the cycle from Figure~\ref{fig:d} is \( \frac{\gamma + \gamma^{3}}{1 - \gamma^{4}} \). For appropriate choices of \( \gamma \), each of the $|S_{2m}|$ cycles can yield a different value for state \( 0 \), hence resulting in exponentially many values. Roughly speaking, this disparity means that a single iteration of PI under average reward could potentially require an exponential number of iterations to cover under discounted reward. In Section~\ref{sec:Disc-reward-UB} we show that HPI, however, can visit at most a subexponential number of cycles (and policies) as a function of \( n \).
\section{Related Work}
\label{sec:related-work}
\citet{mansour1999complexity} established that any run of HPI can take at most $O\left(\frac{k^n}{n}\right)$ iterations, providing the first non-trivial upper bound for the algorithm. \citet{hollanders2012complexity} later improved this bound by a constant factor. To date, the bound of $O\left(\frac{k^n}{n}\right)$ iterations---only a linear improvement over the trivial bound of $k^n$ iterations---remains the tightest known bound for HPI on general MDPs (among those depending solely on $n$ and $k$). 

Experiments suggest that HPI may be much more efficient on MDPs than the current upper bound
%~\citep{mansour1999complexity} would
indicates. Currently-known lower bounds do not rule out this possibility. In an important breakthrough, \citet{fearnley2010exponential} constructed an MDP with a path of length $\exp(\Omega(n))$ for HPI. More recently, \citet{christ2023smoothed} have shown that a form of smoothed complexity~\citep{spielman2004smoothed} for HPI (with an appropriate perturbation model) is super-polynomial, of the form $\exp(\Omega(n^{\frac{1}{3}}))$. However, these results do not settle the complexity of HPI. Notably, in both constructions, $k$, which is the number of actions per state in the MDP, is \textit{not} a free parameter. Rather, $k$ is set to be $\Theta(n)$, where $n$ is the number of states. As yet, the tightest lower bound for HPI on MDPs with constant $k$ is only $\Omega(n)$~\citep{hansen2010lower}. MDPs with $k = 2$ actions induce abstract cubes called acyclic unique sink orientations (AUSOs)~\citep{Szabo+Welzl:2001}. The vertices of the AUSO correspond to policies of the MDP, and oriented edges encode the direction of improvement between neighbours. HPI does have a lower bound of  $\exp(\Omega(n))$ iterations when run on AUSOs~\citep{schurr2005jumping}. However, not all AUSOs are induced by MDPs; it remains unknown if exponentially-long chains are possible for HPI on 2-action MDPs.

\citet{ye2011simplex} provides an upper bound of $\text{poly}\left(n, k, \frac{1}{1 - \gamma}\right)$ iterations for HPI on MDPs; this result was later refined by \citet{hansen2013strategy} and \citet{scherrer2013improved}. Our analysis, which ultimately yields a $\gamma$-independent bound, relies on the following result of \citet{scherrer2013improved}.

\begin{theorem}
\label{thm:scherrer}
Let $M = (S, A, T, R, \gamma)$ be an MDP using discounted reward, with $|S| = n$ and $|A| = k$. On $M$, HPI can visit at most $O\left(\frac{n k}{1 - \gamma} \log \frac{1}{1 - \gamma}\right)$ policies.
\end{theorem}

Although DMDPs are simpler than MDPs, they have themselves challenged theoretical research for many years now~\cite{karp1978characterization,papadimitriou1987complexity,madani2002}
While \citet{papadimitriou1987complexity}, and also \citet{madani2010discounted}, have shown strongly-polynomial algorithms for DMDPs, there is still a gap between upper and lower bounds. \citet{post2013simplex} analyse max-gain Simplex specifically on DMDPs, and show a strongly-polynomial upper bound, which was later improved by \citet{hansen2014}.  

HPI, typically faster in practice than Simplex, is conspicuous by its absence from the results above. \citet{post2013simplex} specifically earmark extending their techniques to HPI as ``a difficult but natural next step''. To the best of our knowledge, the only advance that has subsequently been made is by \citet{goenka2025upper}, who have shown an upper bound of roughly $\text{poly}(n, k) \cdot \left(\frac{k}{e}\right)^{n}$ iterations (when $k$ is large) for HPI on DMDPs. Constraints arising from rewards have not been sufficiently utilised in preceding analyses. It is also to note that lower bound constructions for HPI~\citep{hansen2010lower,fearnley2010exponential,christ2023smoothed} invariably require rewards whose magnitude increases exponentially in $n$---as is seldom the case in practice. Setting out specifically to examine the role of rewards, our work takes a first step by assuming finite bit-sizes. In practice rewards are set by designers; our assumption is somewhat more natural than assumptions made on the transitions (``ADAG" structure by \citet{madani2002policy}; partitioned state space by~\citet{scherrer2013improved}) to obtain tighter bounds for HPI.

The technical core of our analysis is similar to the method used by \citet{grand2024reducing}, though with a different objective. While these authors aim to establish an upper bound on \(\gamma_{\text{bw}}\), the so-called ``Blackwell'' discount factor of an MDP, we focus on tracking the trajectory of a classical algorithm to demonstrate that \(\gamma_{\text{Q}}\) (unknown to the algorithm) constrains its running time nonetheless.

\begin{figure}[b]
  \centering
    \begin{tikzpicture}[scale=1.25,
    >={Stealth[length=6pt,width=7pt]},
    % Specifications
    every edge/.append style={thick},
    vertex/.style={draw, fill=black, circle, minimum size=5pt, inner sep=2pt},
    brown edge/.style={draw=brown},
    teal edge/.style={draw=teal}
    ]

    % vertices
    
    \node[vertex,label={below:\footnotesize 0}] (A) at (0,0) {};
    \node[vertex,label={below:\footnotesize 1}] (B) at (1,0) {};
    \node[vertex,label={below:\footnotesize 2}] (B1) at (2,0) {};
    \node[vertex,label={below:\footnotesize $n-3$}] (C1) at (3.5,0) {};
    \node[vertex,label={below:\footnotesize $n-2$}] (C) at (4.5,0) {};
    \node[vertex,label={below:\footnotesize $n-1$}] (D) at (5.5,0) {};

    % edges (brown)
    \path[->, brown edge] (A) edge node[midway, above, brown] {\scriptsize 0}(B);
    \path[->, brown edge] (B) edge node[midway, above, brown] {\scriptsize 0}(B1);
    \path[->, brown edge] (C1) edge node[midway, above, brown] {\scriptsize 0}(C);
    \path[->, brown edge] (C) edge node[midway, above, brown] {\scriptsize 0}(D);

    % edges (teal)
    \path[->, dashed, teal edge, loop, out=130, in=50, min distance=6mm] (A) edge node[above, teal] {\scriptsize 0} (A);
    \path[->, dashed, teal edge, loop, out=130, in=50, min distance=6mm] (B) edge node[above, teal] {\scriptsize 0} (B);
    \path[->, dashed, teal edge, loop, out=130, in=50, min distance=6mm] (B1) edge node[above, teal] {\scriptsize 0} (B1);
    \path[->, dashed, teal edge, loop, out=130, in=50, min distance=6mm] (C1) edge node[above, teal] {\scriptsize 0} (C1);
    \path[->, dashed, teal edge, loop, out=130, in=50, min distance=6mm] (C) edge node[above, teal] {\scriptsize 0} (C); 
    \path[->, dashed, teal edge, loop, out=130, in=50, min distance=6mm] (D) edge node[above, teal] {\scriptsize 0} (D); 
    \path[->, brown edge, loop, out=80, in=360, min distance=7mm] (D) edge node[right, brown] {\scriptsize 1} (D);     

    \node[draw=none, right of=B1, node distance=1.5cm] (dots) {$\dots$};

    \end{tikzpicture}
\caption{Induced graph of a DMDP with states $S = \{0, 1, \dots, n - 1\}$ and actions $0$ (dashed) and $1$ (solid). Annotations show the reward on each transition. Discount factor $\gamma$ is any element of $(0, 1)$. Policies are encoded as $n$-bit strings: bit $i$ specifies the action taken at state $i - 1$. For starting policy $0^{n}$, the only improving  action is at state $n - 1$. Hence every PI algorithm must proceed from $0^{n}$ to $0^{n - 1}1$. Now, for policy $0^{n - 1}1$, the only improving action is at state $n - 2$. Hence every PI algorithm must proceed from $0^{n - 1}1$ to $0^{n - 2}1^{2}$. The same pattern continues until, after $n$ total iterations, the optimal policy $1^{n}$ is reached.}
\label{fig:dmdp-lb}
\end{figure}
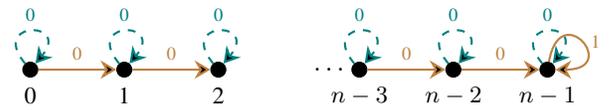

The reader might be curious if an explicit \textit{lower bound} can be shown for HPI in terms of $b$. Recall that with no constraint on $b$ (effectively, $b \to \infty$), the tightest lower bound is currently only $\Omega(n)$~\citep{hansen2010lower}. This lower bound can be achieved up to a constant factor even with $b = 1$ on a DMDP, as shown in Figure~\ref{fig:dmdp-lb}.
Unless a superlinear lower bound is achieved for unrestricted rewards, pursuing a $b$-dependent lower bound does not appear worthwhile.

\section{Pseudosubexponential Upper Bound for Discounted Reward}
\label{sec:Disc-reward-UB}

In this section, we derive our subexponential-in-$n$ upper bound for HPI under discounted reward. To begin, in Section~\ref{sec:discountbound-thresholdgamma}, we define a ``threshold'' discount factor $\gamma_{\text{Q}}$. In Section~\ref{sec:discounted-q-differencesignpolynomial}, we consider and establish the properties of the ``Q-difference sign polynomial'', which is central to our analysis. In Section~\ref{sec:discount-math}, we provide a standalone mathematical result, which we obtain by adapting existing analyses of the roots of polynomials with integer coefficients. This completes the list of ingredients required for our final proof, which we furnish in Section~\ref{sec:discount-finalproof}.

\subsection{Threshold discount factor}
\label{sec:discountbound-thresholdgamma}

Consider DMDP $M = (S, A, T, R, \gamma)$. For policy $\pi: S \to A$; $s \in S$; $a, a^{\prime} \in A$, we define
\begin{multline*}
    \gamma^{\pi}_{s, a, a^{\prime}} \eqdef  
    \inf \bigg\{ \gamma \in [0, 1) \bigg| \ \forall \tau \in (\gamma, 1), \\ 
    \left( Q^{\pi}_{\gamma}(s, a) > Q^{\pi}_{\gamma}(s, a^{\prime}) 
    \implies Q^{\pi}_{\tau}(s, a) > Q^{\pi}_{\tau}(s, a^{\prime}) \right) \bigg\}.
\end{multline*}
In other words, $\gamma^{\pi}_{s, a, a^{\prime}}$ is the greatest lower bound on $\gamma \in [0, 1)$ such that if $a$ has a larger Q-value than $a^{\prime}$ under $\gamma$-discounting, then it also has a larger Q-value under $\tau$-discounting for all $\tau \in (\gamma, 1)$. Although it may not be apparent from the definition, our upcoming working will show that $\gamma^{\pi}_{s, a, a^{\prime}}$ always exists in $[0, 1)$. Consequently, our threshold discount factor
\begin{align}
\label{eqn:gammaq}
\gamma_{\text{Q}} \eqdef \max\limits_{\substack{\pi:S \to A; s \in S; a, a^{\prime} \in A}} \gamma^{\pi}_{s, a, a^{\prime}}   
\end{align}
is also well-defined and guaranteed to lie in $[0, 1)$.

It is insightful to compare $\gamma_{\text{Q}}$ with the ``Blackwell discount factor'' $\gamma_{\text{bw}}$ that is defined by \citet{grand2024reducing}. ``Blackwell-optimality''~\citep{blackwell1962discrete} is an alternative to discounted reward and average reward for defining optimal policies. Like average reward, Blackwell optimality does not require the specification of a discount factor, and may hence be defined for any MDP $M = (S, A, T, R)$. A policy $\pi: S \to A$ is said to be Blackwell-optimal for $M$ if there exists $\gamma \in [0,1)$ such that $\pi$ is discount-optimal for every discount factor $\gamma' \in (\gamma, 1)$. Denote the set of discount-optimal policies for $M$ with discount factor $\gamma$ by $\Pi^{\star}_\gamma$, and the set of Blackwell-optimal policies for $M$ by $\Pi^{\star}_{\text{bw}}$.  \citet{grand2024reducing} define the Blackwell discount factor for $M$ as
\begin{align}
\label{eqn:gamma-bw}
\gamma_{\text{bw}} \eqdef \inf\bigg\{\gamma \in [0,1) \ \bigg | \ \forall \gamma' \in (\gamma,1) \left(\Pi^\star_{\gamma'}=\Pi^\star_{\text{bw}} \right) \bigg\}.
\end{align}
By implication, for every $\gamma \in (\gamma_{\text{bw}}, 1)$, any optimal policy with $\gamma$-discounting is also Blackwell-optimal. 
The significance of this observation arises from a historical context, wherein the computation of Blackwell-optimal policies has not been straightforward. Many procedures for computing Blackwell-optimal policies are relatively complex to implement. In contrast, \citet{grand2024reducing} offer a simple computational recipe when the bit-size of the input MDP is bounded. They derive an explicit upper bound on $\gamma_{\text{bw}}$, which depends on $n$, $k$, as well as the number of bits used to represent the transition probabilities and rewards in $M$. By choosing any discount factor $\gamma$ that exceeds this upper bound, one only has to compute a discount-optimal policy with $\gamma$-discounting in order to obtain a Blackwell-optimal policy.
Although $\gamma_{\text{Q}}$ is syntactically similar to $\gamma_{\text{bw}}$, its intended purpose in this paper is different.
We do not design a new algorithm based on $\gamma_{\text{Q}}$, but rather, highlight its role in determining the running time of an existing, classical algorithm. Correspondingly, 
$\gamma_{\text{Q}}$ is intimately involved with each step performed by HPI, whereas $\gamma_{\text{bw}}$ is only useful to characterise a subset of Blackwell-optimal policies. It can be seen that $\gamma_{\text{Q}}$ is lower-bounded by $\gamma_{\text{bw}}$.

\begin{lemma}
    For every DMDP $(S, A, T, R)$, $\gamma_{\text{bw}} \leq \gamma_\text{Q}$.
\end{lemma}
\begin{proof}
Fix $\gamma_{0} \in (\gamma_\text{Q},1)$ and an arbitrary policy $\overline{\pi} \in \Pi^\star_{\gamma_{0}}$. Since $\overline{\pi}$ is optimal with $\gamma_{0}$-discounting, we have that for each $s \in S; a \in A$,
\begin{equation}\label{eqn:gammaq-gammabw-1}
        Q^{\overline{\pi} }_{\gamma_{0}}(s,\overline{\pi}(s)) \geq Q^{\overline{\pi} }_{\gamma_{0}}(s,a).
\end{equation}
Now suppose there exist $s \in S, a \in A$, and $\gamma_{1} \in (\gamma_{\text{Q}}, \gamma_{0})$ such that $Q^{\overline{\pi}}_{\gamma_{1}}(s, a) > Q^{\overline{\pi}}_{\gamma_{1}}(s, \overline{\pi}(s))$. Since $\gamma_{1} > \gamma_{\text{Q}}$, and therefore $\gamma_{1} > \gamma^{\overline{\pi}}_{s, a, \overline{\pi}(s)}$, and therefore $\gamma_{0} > \gamma^{\overline{\pi}}_{s, a, \overline{\pi}(s)}$, it follows that $Q^{\overline{\pi} }_{\gamma_{0}}(s, a) >Q^{\overline{\pi} }_{\gamma_{0}}(s,\overline{\pi}(s))$---which contradicts \eqref{eqn:gammaq-gammabw-1}. We conclude that for $s \in S, a \in A, \gamma_{1} \in (\gamma_{\text{Q}}, \gamma_{0})$, 
\begin{equation}\label{eqn:gammaq-gammabw-2}
        Q^{\overline{\pi} }_{\gamma_{1}}(s,\overline{\pi}(s)) \geq Q^{\overline{\pi} }_{\gamma_{1}}(s,a).
\end{equation}
In other words, if $\overline{\pi}$ is optimal with $\gamma_{0}$-discounting, it must be optimal with $\gamma_{1}$-discounting for all $\gamma_{1} \in (\gamma_{\text{Q}}, \gamma_{0})$. From \eqref{eqn:gamma-bw}, we observe that $\gamma_{\text{bw}}$ cannot exceed $\gamma_{\text{Q}}$.
\end{proof}
\subsection{Q-difference sign polynomial}
\label{sec:discounted-q-differencesignpolynomial}
As detailed in Section~\ref{sec:PI}, the basis for policy improvement is the comparison of Q-values of different actions at each state. For the working below, fix $\pi: S \to A; s \in S;$ and $a, a^{\prime} \in A$. From \eqref{eqn:discounted-raw} and \eqref{eqn:q-discounted-dmdp}, we have
\begin{align*}
Q^{\pi}_{\gamma}(s, a) &= R(s, a) + \sum_{i = 1}^{p^{\pi}_{T(s, a)}} \gamma^{i} R(P^{\pi}_{T(s, a)}[i]) \\
&\quad + \frac{\gamma^{p^{\pi}_{T(s, a)}}}{1 - \gamma^{c^{\pi}_{T(s, a)}}} \sum_{i = 1}^{c^{\pi}_{T(s, a)}} \gamma^{i} R(C^{\pi}_{T(s, a)}[i]);\\   
Q^{\pi}_{\gamma}(s, a') &= R(s, a') + \sum_{i = 1}^{p^{\pi}_{T(s, a')}} \gamma^{i} R(P^{\pi}_{T(s, a')}[i]) \\
&\quad + \frac{\gamma^{p^{\pi}_{T(s, a')}}}{1 - \gamma^{c^{\pi}_{T(s, a')}}} \sum_{i = 1}^{c^{\pi}_{T(s, a')}} \gamma^{i} R(C^{\pi}_{T(s, a')}[i]).
\end{align*}
Define
\begin{align}
\label{eqn:f-pi}
f^{\pi}_{s, a, a^{\prime}}(\gamma) \eqdef & \left(Q^{\pi}_{\gamma}(s, a) - Q^{\pi}_{\gamma}(s, a^{\prime}) \right) \nonumber \\
& \times \left( 1 - \gamma^{c^{\pi}_{T(s, a)}} \right) \cdot \left( 1 - \gamma^{c^{\pi}_{T(s, a^{\prime})}} \right).
\end{align}
We observe that the sign of $Q^{\pi}_{\gamma}(s, a) -      
Q^{\pi}_{\gamma}(s, a^{\prime})$ is the same as the sign of $f^{\pi}_{s, a, a'}$. Hence, $a$ has a higher Q-value than $a'$ at $s$ under $\pi$ if and only if $f^{\pi}_{s, a, a'}(\gamma) > 0$. By expanding out $f^{\pi}_{s, a, a'}$, we observe that
\begin{align*}
f^{\pi}_{s, a, a'}(\gamma) &= f_{1}(\gamma) + f_{2}(\gamma) + f_{3}(\gamma) + f_{4}(\gamma), \text{ where}\\
f_{1}(\gamma) &= \left( R(s, a) + \sum_{i = 1}^{p^{\pi}_{T(s, a)}} \gamma^{i} R(P^{\pi}_{T(s, a)}[i]) \right) \\
&\quad \times \left( 1 - \gamma^{c^{\pi}_{T(s, a)}} \right) \left( 1 - \gamma^{c^{\pi}_{T(s, a')}} \right),\\
f_{2}(\gamma) &= - \left( R(s, a') + \sum_{i = 1}^{p^{\pi}_{T(s, a')}} \gamma^{i} R(P^{\pi}_{T(s, a')}[i]) \right) \\
&\quad \times \left( 1 - \gamma^{c^{\pi}_{T(s, a)}} \right) \left( 1 - \gamma^{c^{\pi}_{T(s, a')}} \right),\\
f_{3}(\gamma) &= \left( \gamma^{p^{\pi}_{T(s, a)}} \sum_{i = 1}^{c^{\pi}_{T(s, a)}} \gamma^{i} R(C^{\pi}_{T(s, a)}[i]) \right) \\
&\quad \times \left( 1 - \gamma^{c^{\pi}_{T(s, a')}} \right), \text{ and}\\
f_{4}(\gamma) &= - \left( \gamma^{p^{\pi}_{T(s, a')}} \sum_{i = 1}^{c^{\pi}_{T(s, a')}} \gamma^{i} R(C^{\pi}_{T(s, a')}[i]) \right) \\
&\quad \times \left( 1 - \gamma^{c^{\pi}_{T(s, a)}} \right).
\end{align*}

We draw the following observations about $f^{\pi}_{s, a, a^{\prime}}(\gamma)$.

\begin{enumerate}
\item When treated as a function of $\gamma$, $f^{\pi}_{s, a, a^{\prime}}(\gamma)$ is a polynomial with integer coefficients.
\item The degree of $f^{\pi}_{s, a, a^{\prime}}(\gamma)$ cannot exceed the degree of any of its constituents $f_{1}(\gamma)$, $f_{2}(\gamma)$, $f_{3}(\gamma)$, and $f_{4}(\gamma)$. The degree of $f_{1}(\gamma)$ is $\max\{1, p^{\pi}_{T(s, a)}\} + c^{\pi}_{T(s, a)} + c^{\pi}_{T(s, a')}$. Since $p^{\pi}_{T(s, a)} + c^{\pi}_{T(s, a)}$ (and similarly $p^{\pi}_{T(s, a')} + c^{\pi}_{T(s, a')}$) can at most be $n$, the degree of $f_{1}$ is at most $2n + 1$. By a similar argument, the degree of $f_{2}(\gamma)$ is also upper-bounded by $2n + 1$. The degree of $f_{3}(\gamma)$ is $p^{\pi}_{T(s, a)} + c^{\pi}_{T(s, a)} + c^{\pi}_{T(s, a')}$, which is upper-bounded by $2n$, as also is the degree of $f_{4}(\gamma)$ by a similar argument. Thus, the degree of
$f^{\pi}_{s, a, a^{\prime}}(\gamma)$ is at most $2n + 1$.
\item Let the operator $H(\cdot)$ denote the ``height'' of a polynomial, which is the largest absolute value of any coefficient in the polynomial. For example, the height of $2x^{2} - 7x + 3$ is 7. It follows that $H(f^{\pi}_{s, a, a'}) \leq H(f_{1}) + H(f_{2}) + H(f_{3}) + H(f_{4})$. Since each reward lies in $\{0, 1, \dots, 2^{b} - 1\}$, we observe that $H(f_{1})$ and $H(f_{2})$ are at most $4(2^{b} - 1)$, while $H(f_{3})$ and $H(f_{4})$ are at most $2(2^{b} - 1)$. Aggregating, we have $H(f^{\pi}_{s, a, a'}) \leq 12\cdot2^{b}$.
\end{enumerate}
We refer to $f^{\pi}_{s, a, a'}$ as the Q-difference sign polynomial. The three properties established above enable us to upper-bound the roots of $f^{\pi}_{s, a, a'}$ that are smaller than $1$, and this step leads to our final bound for HPI.

\subsection{Distance of certain algebraic numbers from 1}
\label{sec:discount-math}

In this self-contained subsection, we provide an upper bound on the largest root in the interval $(0,1)$ for a general class of polynomials $P$, of degree $n \geq 1$. The notation used in this subsection is aligned with the literature on polynomials (``$P$'' for polynomial, ``$n$'' for degree, ``$a_{(\cdot)}$'' for coefficients, and so on). Any symbols used in this subsection are not to be confused with quantities defined outside it (such as $n$ for the number of states and $a$ for actions). Recall that $H(P)$ is the height of $P$. Below we state our main theorem; the remainder of the subsection provides the proof.

\begin{theorem}
\label{thm:mathmain}
Consider the polynomial:
\[P(x) \eqdef \sum^n_{i=0}{a_i x^i}, \ \ a_i \in \mathbb{Z}. \]
Suppose $\tau < 1$ is a root of $P$: that is, $P(\tau) = 0$. Then $\tau \leq 1 - \frac{1}{U_P}$, where $U_P = O\left (\frac{e^zn^{z+2}}{z^z}H(P)\right)$ and $z \geq 0$ is the multiplicity of root $1$. Further, $z = O \left( \sqrt{n \log H(P)} \right)$.
\end{theorem}

If $\tau$ is a root of $P(x)$, then $1-\tau$ must be a root of $P(1-x)$. Thus to find an upper bound on the root closest to $1$ (from below) for $P(x)$ is equivalent to finding a lower bound on the positive roots of $P(1-x)$.
The general formula for $P(1-x)$ is given below.
    \begin{equation}
        P(1-x) = P(1) + \sum_{j=1}^n{\left[\sum_{i=j}^n{\binom{i}{j}  a_i} \right] (-1)^j x^j}.
    \end{equation}
    
Now, computing a lower bound on the absolute value of roots of a polynomial $J$ is equivalent to computing the reciprocal of an upper bound on the ``reverse polynomial'', formally stated below.
    \begin{proposition}
        Let: \[J(x) = \sum^n_{i=0}{\alpha_i x^i}, \ \alpha_i \in \mathbb{C}. \]
        If $\alpha_n\neq 0$ and $U$ is an upper bound on the absolute value of roots of $J$, then $\frac{1}{U}$ is a lower bound on the absolute values of the roots of $J_r$ where: 
    \[ J_r(x) \eqdef \sum^n_{i=0}{\alpha_{n-i} \cdot x^i}. \]
    \end{proposition}
    We proceed to find an upper bound on the absolute values of the roots of $P_r(1-x)$. We use the following result which was originally proposed by Lagrange, but has also been attributed to Zassenhaus by Knuth~\citep{yap2000fundamental}. 
    \begin{proposition} \label{zassenhaus}
        For a polynomial $J_r$ described above, if $\alpha_0 \neq 0$, an upper bound of the absolute values of the roots of $J_r$ is given by:
        \begin{equation}
        U = 2 \cdot \overset{n}{\underset{s=1}{\text{max}}} \left( \left|\frac{\alpha_s}{\alpha_0} \right| ^{1/s} \right).
        \end{equation}
    \end{proposition}
    Applying Proposition~\ref{zassenhaus} on $P_r(1-x)$ leads to two cases. \\ \\
    \textit{Case 1}: Suppose $P(1)\neq 0$. Since $P$ has integer coefficients, it follows that $|P(1)| \geq 1$. From the proposition,
     the upper bound is given by:     \begin{align*}
     U_P &= 2 \cdot \overset{n}{\underset{s=1}{\text{max}}} \left( \left|\frac{\sum_{i=s}^n{\binom{i}{s} a_i}}{P(1)} \right| ^{\frac{1}{s}} \right) \\
     &\leq 2 \cdot \overset{n}{\underset{s=1}{\text{max}}} \left( \left| \sum_{i=s}^n { \binom{i}{s} a_i }\right|^{\frac{1}{s}}  \right)\\ &\leq 2 \cdot \overset{n}{\underset{s=1}{\text{max}}} \left( \left| \sum_{i=s}^n {\binom{i}{s}}\right|^{1/s} \left(H(P)\right)^{\frac{1}{s}}   \right).
    \end{align*}
     Now $\sum_{i=s}^n { \binom{i}{s}  } = \binom{n+1}{s+1}$. 
     And since $H(P) \geq 1$, the maximum occurs at $s=1$, giving
     \begin{equation} \label{UB-case 1}
          U_P \leq 2 \cdot \binom{n+1}{2} H(P) = n(n+1)\!\cdot\!H(P).
     \end{equation}
    \textit{Case 2:} Suppose $P(1)=0$: that is, $1$ is a root of $P$. If the multiplicity of the root $1$ is $z \geq 1$, we have 
    \begin{align*}
    P(x) &= (x-1)^{z}D(x), \text{ where}\\
    D(x) &= \sum^{n-z}_{i=0}{d_i x^i}, \ \ d_i \in \mathbb{Z}, D(1) \neq 0.
    \end{align*}
    We provide the formula for the $i$-th term of $D$ below. 
    \[ d_i = \sum_{j=i}^{n-z}{\binom{n-j-1}{n-j-z} a_{n-j+i}}.\]
    Now, since
    \[\sum_{j=0}^{n-z}{\binom{n-j-1}{n-j-z} }= \binom{n}{z},\]
    for all $i$, we get 
    \[ |d_i| \leq \binom{n}{z} H(P).\]
    Since $D(1)\neq 0$, we can use \eqref{UB-case 1} (from case 1) to get a root upper bound for $D(x)$:
    \begin{align}
        U_D & \leq n(n+1)H(D) \notag \leq n(n+1)\binom{n}{z} H(P) \notag\\
        % & \leq n(n+1){n \choose z} H \notag \\ 
        & \leq n(n+1)\left(\frac{en}{z}\right)^z H(P) \notag \leq O\left (\frac{e^zn^{z+2}} {z^z}H(P)\right).\\
        \label{UB-case2}
    \end{align}
Note that since $P(x) = (x - 1)^{z}D(x)$, $U_{P} = U_{D}$. Although the bound from \eqref{UB-case 1} (case 1) shows only polynomial growth with $n$, the one from \eqref{UB-case2} depends on the number of roots $z$ that $P$ can possibly have. The following result from \citet{borwein1999littlewood} lets us upper-bound $z$. 
% https://math.stackexchange.com/questions/1352338/proof-for-the-upper-bound-and-lower-bound-for-binomial-coefficients

\begin{theorem}\label{Borwein}
    \citep{borwein1999littlewood} There is an absolute constant $c>0$ such that every polynomial $p$ of the form
\[ p(x) = \sum_{j=0}^n {a_jx^j}, \ \ |a_j|\leq 1, a_j \in \mathbb{C} \]
has at most 
\[c(n(1-\log |a_0|))^{1/2}\]
zeros at $1$.
\end{theorem}

By Theorem \ref{Borwein}, an upper bound on the number of zeros at $1$ of the polynomial $\frac{P(x)}{H(P)}$ and therefore $P(x)$ is given by: \[c(n(1 + \log H(P)))^{1/2}=O \left( \sqrt{n \log H(P)} \right). \]

In summary, we have shown that for every $\tau < 1$ that is a root of  $P$: 
\begin{multline*}
\tau \leq 1 - \frac{1}{U_{P}}, \text{ where } \ U_P\leq O\left (\frac{e^zn^{z+2}} {z^z}H(P)\right), \text{ and} \\ z \leq O \left( \sqrt{n \log H(P)} \right).
\end{multline*}
This concludes the proof of Theorem~\ref{thm:mathmain}.

\subsection{Proof of Theorem~\ref{thm:discountedreward}}
\label{sec:discount-finalproof}

We are now ready with our concluding arguments for proving Theorem\ref{thm:discountedreward}. In Section~\ref{sec:discounted-q-differencesignpolynomial}, we showed that for every $\pi: S \to A; s \in S; a, a' \in A$, the polynomial $f^{\pi}_{s, a, a'}(\gamma)$ has integer coefficients, degree at most $2n + 1 \leq 3n$, and height at most $O(2^{b})$. Since $f^{\pi}_{s, a, a'}(\gamma)$ does not change sign in $(\gamma^{\pi}_{s, a, a'}, 1)$, we see that that $\gamma^{\pi}_{s, a, a'}$ is the largest root of $f^{\pi}_{s, a, a'}(\cdot)$ that is smaller than $1$ (but clipped at $0$). Applying Theorem~\ref{thm:mathmain}, we observe that:
\begin{equation*}
\gamma^{\pi}_{s, a, a'} \leq 1 - \frac{1}{U}, \ U = O\left( \frac{e^{z}(3n)^{z+2}}{z^z} 2^b \right), \ z = O\left( \sqrt{nb} \right).
\end{equation*}
Since $\left(\frac{ne}{z}\right)^z$ is strictly increasing for $z\leq n$,
we get:
\[\log U \leq O\left( \sqrt{nb} \log{ \frac{n}{b}} + b \right).\]

Now from \eqref{eqn:gammaq}, we have $$\gamma_{\text{Q}} \leq 1 - \frac{1}{U}, \text{ or equivalently, } \frac{1}{1 - \gamma_{\text{Q}}} \leq U.$$

For $\gamma \in [0, \gamma_{\text{Q}})$, our upper bound in Theorem~\ref{thm:discountedreward} is trivially valid due to the $\gamma$-dependent upper bound of \citet{scherrer2013improved}, which we have rephrased in Theorem~\ref{thm:scherrer}. Now consider $\gamma \in (\gamma_{\text{Q}}, 1)$. At any iteration of HPI, with policy $\pi$ and state $s$, if $Q^{\pi}_{\gamma}(s, a) \geq Q^{\pi}_{\gamma}(s, a')$ for $a, a' \in A$, then $Q^{\pi}_{\gamma'}(s, a) \geq Q^{\pi}_{\gamma'}(s, a')$ for all $\gamma^\prime \in (\gamma_{\text{Q}}, 1)$. Since HPI always picks an action maximising the Q-value, starting from any policy, it would visit an identical sequence of policies for any $\gamma^{\prime} \in (\gamma_{\text{Q}}, 1)$. In particular consider $\gamma'$ that is arbitrarily close to $\gamma_{\text{Q}}$. By Theorem~\ref{thm:scherrer}, the number of iterations taken by HPI on $(S, A, T, R, \gamma^{\prime})$ would be
$O\left(\frac{nk}{1 - \gamma_{\text{Q}}} \log \frac{1}{1 - \gamma_{\text{Q}}}\right)$, which we have just shown to be $$n k \left(\sqrt{nb} \log\frac{n}{b} + b\right) \exp\left(O\left(\sqrt{nb} \log\frac{n}{b} + b\right)\right),$$ matching the claim in Theorem~\ref{thm:discountedreward}.

\section{Conclusion}
\label{sec:conclusion}

HPI~\citep{howard1960dynamic} has, over several decades, cemented its position as a method of choice among practitioners for solving MDPs. However, known upper bounds for HPI that hold independently of the discount factor are exponentially separated from known lower bounds---not only for MDPs, but even for the restricted subclass of DMDPs. In this paper, we give the first subexponential upper bound for HPI on DMDPs with discounted reward, contingent on an assumption about the bit-size of rewards. This assumption is reasonable in practice, as reward bit-sizes are almost always constant. Theoretically, our bound remains valid and significant even when there are exactly two reward values of arbitrary size. En route to this result, we also show a pseudopolynomial upper bound for HPI on DMDPs under average reward.

We employ a novel analytical technique showing that when the discount factor is ``large,'' every choice made by HPI matches the one it would make with a smaller, ``threshold'' discount factor. We then upper-bound this threshold by proving a result on polynomials with integer coefficients. These steps let us piggyback on a $\gamma$-dependent upper bound from \citet{ye2011simplex} and \citet{scherrer2013improved}, by plugging in the threshold discount factor.
% We develop a novel technique showing that when the discount factor is “large,” each choice made by HPI matches the one it would make with a smaller “threshold” discount factor. We then upper-bound this threshold by proving a result on polynomials with integer coefficients. These steps let us leverage the $\gamma$-dependent upper bound from \citet{ye2011simplex} and \citet{scherrer2013improved} by plugging in the threshold value.

From the related work of \citet{grand2024reducing}, it appears unlikely that the same technique can show a subexponential bound for HPI on general MDPs. However, it would be interesting to investigate whether intermediate classes of MDPs (which interpolate in some manner between DMDPs and MDPs) can benefit from our approach. It may also be possible to improve our subexponential upper bound by exploiting more properties of the Q-difference sign polynomial. On a related note, it is worth exploring if the dependence on $b$, which is currently exponential in both our upper bounds, can be improved.

% Based on \citet{grand2024reducing}, it appears unlikely that the same technique yields a subexponential bound for HPI on general MDPs. However, it would be interesting to explore whether intermediate classes of MDPs (interpolating between DMDPs and MDPs) can benefit from our approach. It may also be possible to tighten our subexponential bound by exploiting more properties of the Q-difference sign polynomial. Finally, it is worth exploring if the dependence on $b$, currently exponential in both our bounds, can be improved.
% \section*{Ethical Statement}
% The authors do not foresee any significant ethical or societal consequences arising from this work.
\section*{Acknowledgements}
The authors
thank Sundar Vishwanathan, Supratik Chakraborty, Akash 
Kumar, and Rohit Gurjar for providing useful comments.
\bibliography{References}
\clearpage
\appendix
\onecolumn
\section{Analysis under Average Reward}
\label{app:averagereward}
The ``average cost''  criterion (cost is the negative of reward)
is a commonly used method for optimising stochastic dynamical systems over an infinite time horizon.
Solving DMDPs under the average cost criterion is equivalent to finding minimum mean cost cycles (MMCC) in the induced graph. \citet{karp1978characterization} provided a polynomial time algorithm to the MMCC problem. However, the complexity of Howard's Policy Iteration (HPI) in this context remains an open problem, even though, in practice, HPI is typically more efficient than Karp's algorithm~\cite{dasdan2004experimental}. On the other hand, \citet{hansen2010lower} established a lower bound of $\Omega(n^2)$ for HPI when the graph has $\Theta(n^2)$ edges. We provide a polynomial upper bound for all policy iteration algorithms under the assumption that the bit-size of the rewards 
is constant.

In this section we describe the notion of average reward and present the proof of Proposition~\ref{prop:averagereward}.

\subsection{Preliminaries}
The average reward values defined in equations (\ref{eqn:average-gain}) and (\ref{eqn:average-bias}) satisfy the pair of Bellman Equations given by:

\begin{align}
\label{eqn:gain-bellman}
V^\pi_{\text{g}}(s) &= \sum_{s' \in S} T(s,\pi(s),s')V^\pi_{\text{g}}(s'),\\
\label{eqn:bias-bellman}
V^\pi_{\text{b}}(s) &= R(s,\pi(s)) - V^\pi_{\text{g}}(s) +  \sum_{s' \in S} T(s,\pi(s),s')V^\pi_{\text{b}}(s').
\end{align}

Note that the system of equations above is underdetermined, meaning that solutions to equations~\eqref{eqn:gain-bellman} and~\eqref{eqn:bias-bellman} are not unique. 
% A common approach to resolve this is to select a state \( s \) in each recurrent class and set \( V_b(s) = 0 \) (see below).
A standard procedure for ensuring a unique bias function is to require that the long-run average bias over time is zero, known as the canonical bias (see~\citet{puterman2014markov}), which is consistent with the definition based on expectations. Another common approach---which we adopt---is to select a state \(s\) in each recurrent class and set \(V_b(s) = 0\). 
% Since every bias function must satisfy the linear equation~\eqref{eqn:bias-bellman}, any two valid bias functions within a recurrent class $\Phi$ differ only by a constant offset---i.e., \(V_{b_1}(s) = V_{b_2}(s) + C\) for all $s \in S_\Phi$. As a result, the analysis we present below holds uniformly for all such solutions.
Additionally, in multichain average reward MDPs, computing the optimal gain and identifying optimal policies require two optimality equations, given by:
\begin{align}
    \label{eqn:gain-opt-eqn}
    \max\limits_{\substack{ a \in A}}\left\{ V_{\text{g}}(s) - \sum_{s' \in S} T(s,a,s')V_{\text{g}}(s') \right\} &= 0,\\
    \label{eqn:bias-opt-eqn}
    \max\limits_{\substack{ a \in A^\prime}} \left\{ -V_{\text{b}}(s) + R(s,a)  - V_{\text{g}}(s) +  \sum_{s' \in S} T(s,a,s')V_{\text{b}}(s') \right\} &= 0,
\end{align}
where $A' = \bigg \{a' \in A \ : \ V_{\text{g}}(s) - \sum_{s' \in S} T(s,a',s')V_{\text{g}}(s') = 0 \bigg \}$.
A solution to the optimality equations always exists for finite MDPs~\cite{puterman2014markov}.

\textbf{Policy iteration} is analogous to the discounted case, except that it maintains and updates a pair of value functions $(V_{\text{g}},V_{\text{b}})$, in a lexicographical order. The following pseudocode outlines this process. 

\begin{center}
\framebox[0.8\textwidth]{
\parbox{0.75\textwidth}{
\underline{PI starting from policy $\pi$ under average reward} \\\\
\underline{\textit{Policy Evaluation:}} \\ \\
1. Obtain $V^\pi_{\text{g}}$ and $V^\pi_{\text{b}}$ which satisfy~\eqref{eqn:gain-bellman},~\eqref{eqn:bias-bellman}.  \\ \\
\underline{\textit{Policy Improvement:}} \\ \\
2.  Let:
\[ J^\pi_\text{g} = \bigg \{ (s,a) \ \bigg | \ s \in S,a\in A, \ \sum_{s' \in S}{ T(s,a,s')V^\pi_{\text{g}}(s')} > V^\pi_\text{g}(s) \bigg \}. \]
\\

3. If $J^\pi_\text{g} \neq \emptyset$, pick $I_\text{g} \subseteq J^\pi_\text{g} \ \left(I_\text{g}\neq \emptyset\right)$ and let $\pi'(\overline{s}) \gets \overline{a}$ for $(\overline{s},\overline{a}) \in I_\text{g}$ and $\pi'(s) \gets \pi(s)$ for the remaining states $s$. Set $\pi \gets \pi'$ and go to 1. Else go to 4. \\ \\

4. Let 
\[ J^\pi_\text{b} = \bigg \{ (s,a) \ \bigg | \ s \in S,a\in A, \  R(s,a) + \sum_{s' \in S}{T(s,a,s')V^\pi_{\text{b}}(s')} > V^\pi_\text{b}(s) \bigg \}. \]

5. If $J^\pi_\text{b} \neq \emptyset$, pick $I_\text{b} \subseteq J^\pi_\text{b} \ \left(I_\text{b}\neq \emptyset\right)$ and let $\pi'(\overline{s}) \gets \overline{a}$ for $(\overline{s},\overline{a}) \in I_\text{b}$ and $\pi'(s) \gets \pi(s)$ for the remaining states $s$. Set $\pi \gets \pi'$ and go to 1. Else declare $\pi$ to be optimal and terminate. \\ \\
}
}
\end{center}

A subset of states $S^{\prime} \subseteq S$ is called a \textit{recurrent class} under $\pi$ if for each pair $s, s^{\prime} \in S$, there is a non-zero probability of reaching $s^{\prime}$ from $s$ in fewer than $n$ steps by following $\pi$, and moreover, no state in $S \setminus S^{\prime}$ is reachable from any state in $S^{\prime}$ under $\pi$. The system of equations \eqref{eqn:gain-bellman} and \eqref{eqn:bias-bellman} becomes uniquely determined if any one state in each recurrent class of $\pi$ is allotted any arbitrary bias. Assuming an indexing of states---concretely, take $S = \{0, 1, 2, \dots, n - 1\}$---we adopt the convention of setting $V^{\pi}_{\text{b}}(s) = 0$ for every state $s$ that is in a recurrent class and has the smallest index in that class~\citep{hansen2013strategy}. Since every bias function must satisfy the linear equation~\eqref{eqn:bias-bellman}, any two valid bias functions within a recurrent class $S' \subseteq S$ differ only by a constant offset---i.e., \(V_{b_1}(s) = V_{b_2}(s) + C\) for all $s \in S'$. As a result, the analysis we present below holds uniformly for all such solutions.

\subsubsection{DMDPs.}
For DMDP $M = (S, A, T, R)$ using average reward and policy $\pi: S \to A$, the gain of each state $s \in S$ becomes the average of the rewards on $C^{\pi}_{s}$, which indeed constitutes a recurrent class.
\begin{align}
\label{eqn:gain-raw}
V^{\pi}_{\text{g}}(s) = \frac{1}{c^{\pi}_{s}} \sum_{i = 1}^{c^{\pi}_{s}} R(c^{\pi}_{s}[i]).
\end{align}
The bias $V^{\pi}_{{\text{b}}}(s)$ distinguishes between states that reach the same recurrent class, and is given by
\begin{align}
\label{eqn:bias-raw}
V^{\pi}_{{\text{b}}}(s) = \begin{cases}
0  & \text{if } s \text{ has the smallest index in } C^{\pi}_{s}, \\
R(s, \pi(s)) - V^{\pi}_{{\text{g}}}(s) + V^{\pi}_{\text{b}}(T(s, \pi(s))) & \text{otherwise.}\\
\end{cases}
\end{align}

\subsection{Pseudopolynomial Upper Bound for Average Reward}
We now present a proof of Proposition~\ref{prop:averagereward} which yields the pseudopolynomial bound.

\subsubsection{Proof of Proposition~\ref{prop:averagereward}}

Recall from \eqref{eqn:gain-raw} that the gain of state $s \in S$ under policy $\pi: S \to A$ is equal to $$V^{\pi}_{\text{g}}(s) = \frac{\sum_{i = 1}^{c^{\pi}_{s}} R(c^{\pi}_{s}[i])}{c^{\pi}_{s}}.$$ The numerator is a sum of $c^{\pi}_{s}$ rewards, each an integer in $\{0, 1, \dots, n \! \cdot \! (2^{b} - 1)\}$, while the denominator is an integer in $\{1, 2, \dots, n\}$. Thus, $V^{\pi}_{\text{g}}(s)$ can take at most $T_{1} \eqdef n^2 \cdot 2^{b}$ values.

Similarly, the bias $V^{\pi}_{\text{b}}(s)$ is seen from \eqref{eqn:bias-raw} to be the sum of some $i$ terms of the form $R(s_{i}, \pi(s_{i})) - V^{\pi}_{\text{g}}(s_{i})$, where $0 \leq i \leq n - 1$, and $s_{i}$ is the $i$-th state visited after starting from $s$. The sum of the $i$ reward terms has to be an integer in $\{0, 1, \dots, (n - 1) \cdot (2^{b} - 1)\}$. Since the states $s_{i}$ that are visited from $s$ by following $\pi$ all have the same gain $V^{\pi}_{\text{g}}(s)$, the amount subtracted is from the set $\{0, V^{\pi}_{\text{g}}(s), 2 V^{\pi}_{\text{g}}(s), \dots, (n - 1)V^{\pi}_{\text{g}}(s)\}$. Hence, if $V^{\pi}_{\text{g}}(s)$ is fixed, the number of possible values $V^{\pi}_{\text{b}}(s)$ can take is at most $T_{2} \eqdef n \cdot 2^{b} \cdot n$. Therefore, there are at most \( T_1 \! \cdot \! T_2 \) unique gain-bias pairs possible for any state \( s \). 

 By the Policy Improvement Theorem~\cite{howard1960dynamic}, in each PI improvement step: (1) the gain of each state either remains the same or increases; (2) if gains are unchanged, the bias either remains the same or increases; and (3) at least one state’s gain or bias strictly increases. Thus, any PI algorithm, regardless of the variant, follows a strictly monotonic trajectory starting from the initial policy, with each iteration requiring at least one state to adopt a new gain-bias pair. Consequently, the total number of iterations is bounded by \( n \! \cdot \! T_1 \! \cdot \! T_2 = O(n^5 \cdot 4^b) \).

\end{document}